\documentclass[journal]{IEEEtran}
\usepackage{graphicx}
\usepackage{color}
\usepackage[numbers,sort&compress]{natbib}
\usepackage{algorithm,algpseudocode,algorithmicx}
\usepackage{amsmath,latexsym,amssymb,amsthm,array,amsfonts,algorithm,algpseudocode,booktabs,graphicx,subfigure,multirow,cuted,stfloats}
\ifCLASSINFOpdf
\else
\fi
\begin{document}
%
\title{Augmented Transformers with Adaptive n-grams Embedding for Multilingual Scene Text Recognition} 
%
%
%

\author{Xueming~Yan,
        Zhihang~Fang,
        and ~Yaochu~Jin,~\IEEEmembership{Fellow,~IEEE,}
\thanks{ X. Yan is with the School of Information Science and Technology, Guangdong University of Foreign Studies, Guangzhou 510000, China. She is also with the Faculty of Technology, Bielefeld University, 33619 Bielefeld, Germany. Email: yanxm@gdufs.edu.cn}
\thanks{Z. Fang is with the School of information science and engineering, East China University of Science and Technology, Shanghai 200237, China. Email: yuzaifzh98@163.com.}
\thanks{Y. Jin is with the Faculty of Technology, Bielefeld University, 33619 Bielefeld, Germany. E-mail: yaochu.jin@uni-bielefeld.de. (\textit{Corresponding author})}

\thanks{Manuscript received xxxx, 2023; revised xxxx, 2023.}}

%
%

\markboth{Journal of \LaTeX\ Class Files,~Vol.~xx, No.~x, August~20xx}%
{Shell \MakeLowercase{\textit{et al.}}: Bare Demo of IEEEtran.cls for IEEE Journals}
%



\maketitle

\begin{abstract}
While vision transformers have been highly successful in improving the performance in image-based tasks, not much work has been reported on applying transformers to multilingual scene text recognition due to the complexities in the visual appearance of multilingual texts. To fill the gap, this paper proposes an augmented transformer architecture with n-grams embedding and cross-language rectification (TANGER). TANGER consists of a primary transformer with single patch embeddings of visual images, and a supplementary transformer with adaptive n-grams embeddings that aims to flexibly explore the potential correlations between neighbouring visual patches, which is essential for feature extraction from multilingual scene texts. Cross-language rectification is achieved with a loss function that takes into account both language identification and contextual coherence scoring. Extensive comparative studies are conducted on four widely used benchmark datasets as well as a new multilingual scene text dataset containing Indonesian, English, and Chinese collected from tourism scenes in Indonesia. Our experimental results demonstrate that TANGER is considerably better compared to the state-of-the-art, especially in handling complex multilingual scene texts.
\end{abstract}

\begin{IEEEkeywords}
Evolutionary optimization, neural architecture search, node inheritance, attention mechanism, convolutional neural networks.
\end{IEEEkeywords}

%
\IEEEpeerreviewmaketitle

\section{Introduction}
\label{sec:intro}
 
Transformers have achieved tremendous success in computer vision \cite{han2020survey} in a variety of image-based tasks, including object detection \cite{zhou2017scene}, semantic segmentation \cite{liu2021swin}, and image recognition \cite{dosovitskiy2020image,atienza2021vision}. Since transformers were originally developed for natural language processing \cite{devlin2018bert}, demanding efforts are required to design and train transformers for vision tasks \cite{chen2020generative, feng2020scene}. One attractive idea \cite{dosovitskiy2020image} is to apply a pure transformer, called ViT, directly to sequences of image patches, significantly reducing the required computational resources. 

\begin{figure}
  \begin{center}
   \includegraphics[width=0.9 \columnwidth]{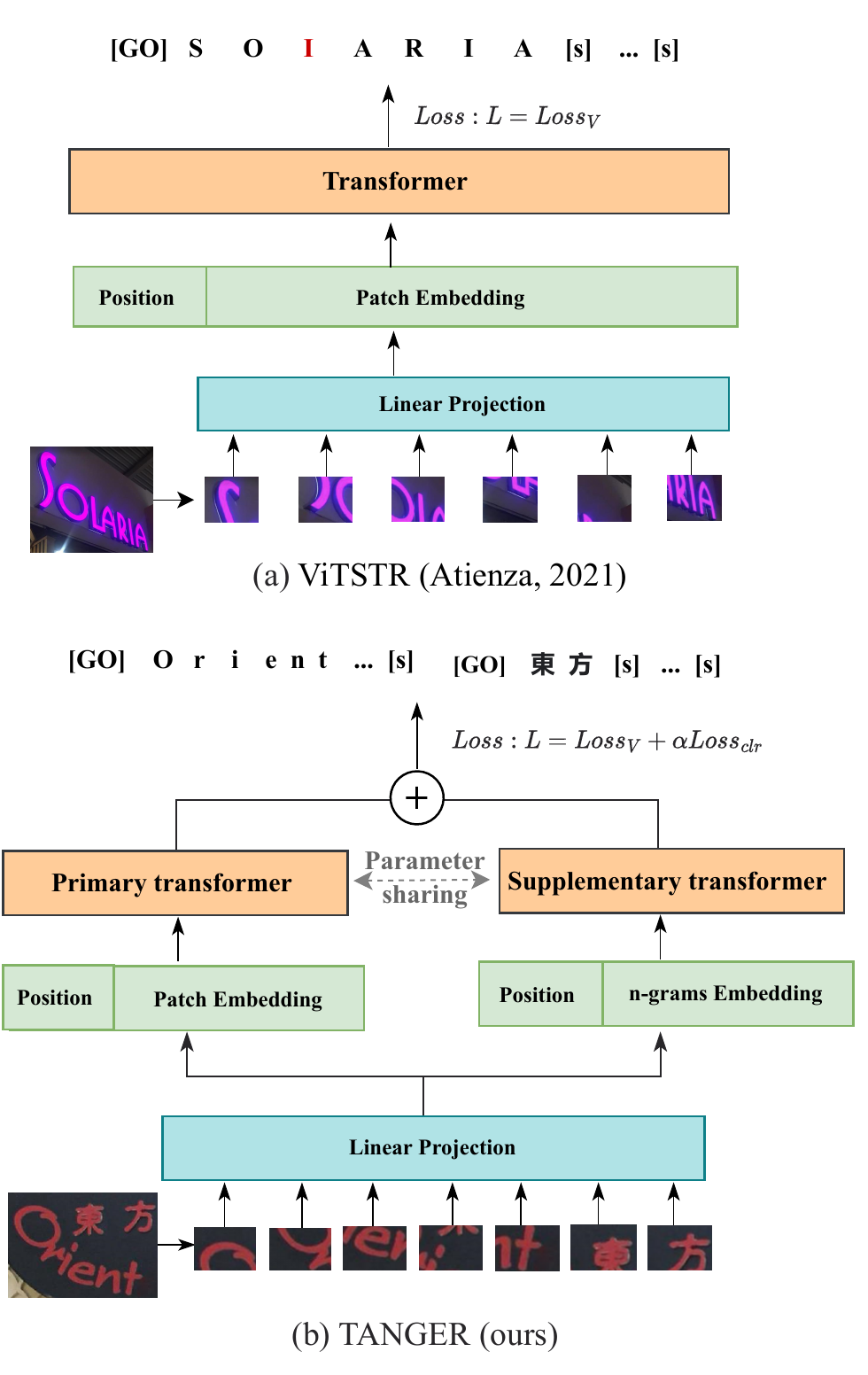}
\end{center}
   \caption{\textbf{Pure transformers for scene text recognition.} (a) ViTSTR (Atienza, 2021), an existing transformer for monolingual scene text recognition.  (b) TANGER: The proposed augmented  transformer architecture with adaptive n-grams embedding and cross-language rectification for multilingual scene text recognition, where $Loss_{V}$ means the loss for vision model and $Loss_{clr}$ means the loss of cross-language rectification.
   }
   \label{fig1}
\end{figure}

Scene text recognition (STR), one class of important and challenging tasks involving both image and text features, aims to identify texts in natural scenes such as object labels, street and road signs \cite{zhu2016scene}. On the basis of ViT, an image patch-based transformer for STR, called ViTSTR, has been proposed in \cite{atienza2021vision} to achieve an optimal trade-off between performance and speed. However, as illustrated in Fig.~\ref{fig1}(a), ViTSTR works directly on sequences of single image patches without considering their associated neighbors, which may be inadequate in dealing with more complex scenes such as multilingual scene texts. Despite their potential competitive performances for STR tasks, not much attention has been paid to vision transformers to address the specific challenges encountered in multilingual scene text recognition \cite{buvsta2018e2e}. 

Multilingual scene text recognition is particularly challenging in that it involves multi-scale, multi-orientation, and low-resolution images containing multiple languages. For instance, the scene texts in Fig.~\ref{fig1}(b) contain English and Chinese, which have different sizes and orientations, making it hard for existing transformers that use single image patches to efficiently recognize multilingual scene texts. 


To tackle the above challenges, we propose an augmented Transformer architecture with Adaptive N-Grams Embeddings and cross-language Rectification (TANGER) for multilingual scene text recognition. Our key hypothesis here is that adaptive patch-based n-grams embeddings can allow TANGER to deal with complex multilingual scene characters more flexibly. In addition to the single image patches for the primary vision transformer, the supplementary pyramid transformer takes various representations of the neighboring image patches as the input. The primary and supplementary vision transformers jointly learn the representation of multilingual scene text features by sharing their parameters, reducing the computational complexity of the proposed model. To further improve the performance for multilingual scene text recognition, a loss function for cross-language rectification is designed for text prediction in the presence of multiple languages by considering the language category information as well as the coherence of a sequence of characters or words. 

We evaluate the proposed TANGER by comparing its performance with the state-of-the-art approaches on three public multilingual benchmark datasets including E2E-MLT \cite{buvsta2018e2e}, CTW1500 \cite{yuliang2017detecting}, and RCTW17 \cite{shi2017icdar2017}, and one monolingual dataset, Totaltext \cite{ch2017total}. In addition, comparative experiments are carried out on a new multilingual database we collected from tourism scenes in Indonesia, called TsiText, which consists of Indonesian, English, and Chinese. Our experimental results demonstrate that TANGER achieves state-of-the-art results on all multilingual and monolingual scene text recognition tasks considered in this work. 

The key contributions of this work are summarized as follows:
\begin{itemize}
    \item We propose an augmented transformer architecture (TANGER) for multilingual scene text recognition by integrating a primary vision transformer with a supplementary pyramid transformer with n-grams embeddings. To our knowledge, this is the first pure vision transformer-based method for multilingual scene text recognition. 
    \item An adaptive n-grams embedding method is designed to more flexibly extract key text features in locally related neighboring visual patches. The adaptive n-grams embedding can determine the optimal number of grams for different input image patches, which is highly beneficial for complex scent text recognition. 
    \item To further enhance TANGER's capability of handling multilingual scene texts, we design a cross-language rectification loss function to account for language identification errors and text coherence.  
\end{itemize}
\begin{figure*}
  \centering
   \includegraphics[width=0.8\linewidth]{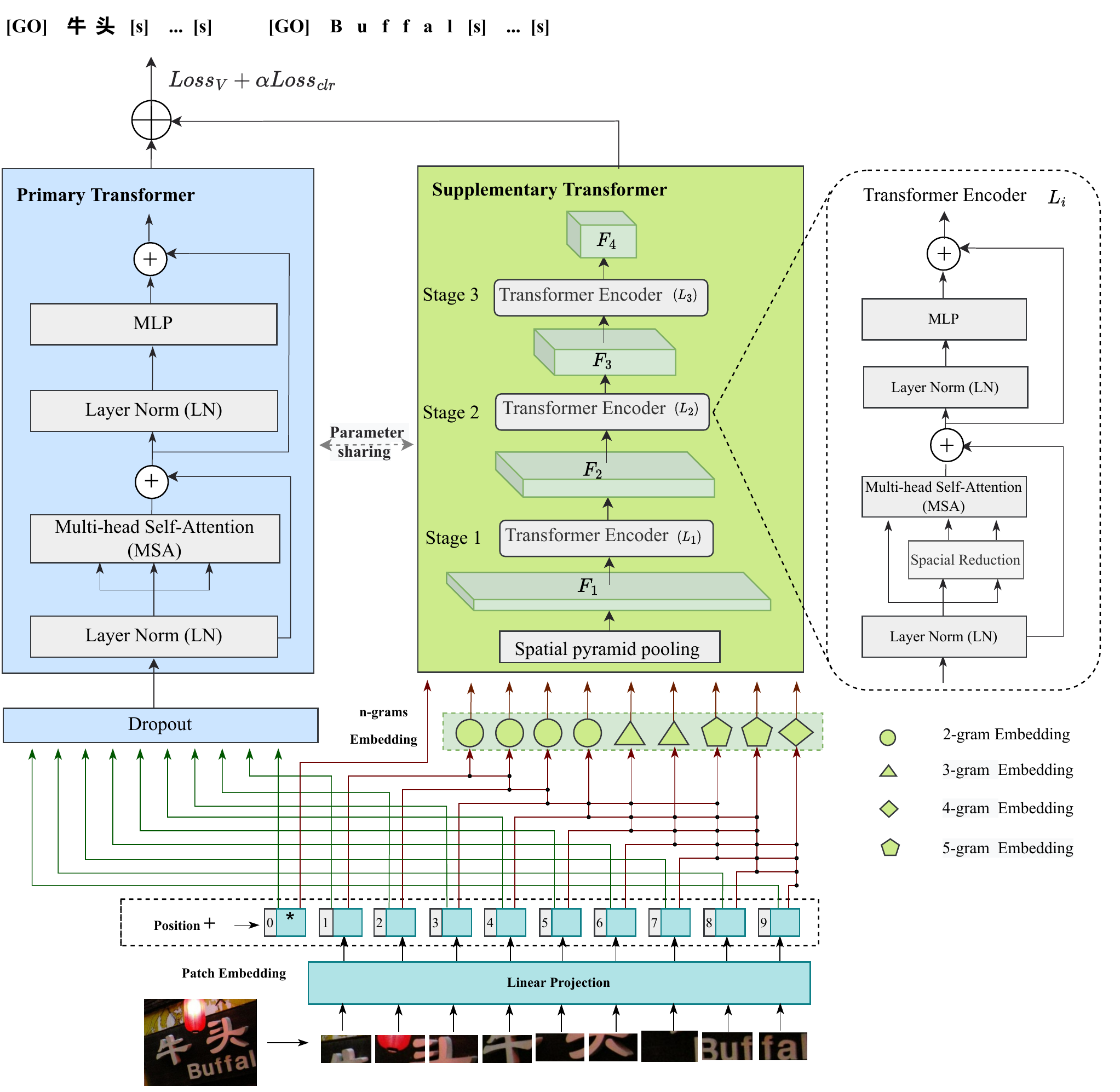}
   \caption{\textbf{Overall architecture of TANGER}. The primary and supplementary transformers are employed to deal with regular and  n-grams embedding visual information, respectively, where $F_{1}$, $F_{2}$,
   $F_{3}$ and $F_{4}$ are the feature maps. 
   }
   \label{fig2}
\end{figure*}
 
\section{Related Work}
\subsection{Scene Text Recognition}
Existing work on STR \cite{xie2019aggregation} can be categorized into language-free and language-based approaches \cite{zhu2016scene}. Language-free methods do not consider the linguistic information between characters and focus on the visual textures for recognition \cite{tounsi2018multilingual}. Yao \textit{et al.} \cite{yao2014strokelets} combine various feature descriptors with localized individual characters, and present a multi-scale representation for STR. Shi \textit{et al.} \cite{shi2016end} propose a unified neural network architecture framework for STR that integrates feature extraction, sequence modeling, and transcription, considering STR as a pixel-wise classification task. Language-based methods focus on the language mode with different structures, or consider the relationship between vision and language. For example, attention-based methods \cite{li2019show,zhan2019esir} are adopted for STR by following an end-to-end neural network model in recurrent neural networks \cite{shi2018aster}. Fang \textit{et al.} \cite{fang2018attention} propose a text recognizer based on convolutional neural networks (CNNs) by fusing visual and language-based methods to boost the recognition performance.

Little work on multilingual scene text recognition has been reported with only a few exceptions. Busta \textit{et al.} \cite{buvsta2018e2e} design probably the first method, called E2E-MLT, for multilingual scene text recognition on the basis of a single fully convolutional network, which is demonstrated to be competitive even on rotated and vertical text instances. An E2E approach to script identification with different recognition heads, called Multiplexed Multilingual Mask TextSpotter, was proposed in \cite{huang2021multiplexed}, which can support the removal of existing or inclusion of new languages. 

However, due to the diversity of scene texts, complexity of the background, large amounts of uncertainty \cite{qiao2020seed}, and different sizes of different language scripts, existing methods fail to work effectively on multilingual scene text recognition tasks. 

\subsection{Vision Transformers for STR}
To benefit from the generated visual features following linguistic rules, increased research interests have been dedicated to using transformers for STR recently \cite{lyu20192d, dosovitskiy2020image,na2021multi}, where the encoder extracts visual features and the decoder predicts characters in images. Owing to their parallel self-attention and prediction mechanisms, transformers can overcome the difficulties of sequential inference with diverse scene text features to some extent \cite{han2020survey}.

Sporadic research efforts on adapting transformers to addressing various challenges in monolingual STR have been reported. For instance, to deal with images with different resolutions, Raisi  \textit{et al.} \cite{raisi2020} develop a transformer-based architecture for recognizing texts in images by using a 2D positional encoder so that the spatial information the features can be preserved. Biten \textit{et al.} \cite{biten2022latr} propose a layer-aware transformer with a pre-training scheme on the basis of text and spatial cues only and show that it works well on scanned documents to handle multimodality in scene text visual question answering. Based on ViT \cite{dosovitskiy2020image}, Tan \textit{et al.} \cite{tan2022pure} propose a mixture experts of pure transformers for processing different resolutions for scene text recognition. Atienza \cite{atienza2021vision} presents a new transformer, called ViTSTR, for scent text recognition, which only uses the encoder architecture and emphasizes the balance between the performance and computational efficiency. Recently, Wang \textit{et al.} \cite{wang2022petr} explore the linguistic information between the local visual patches and propose transformer-based language rectification modules for optimizing word length and guiding rectification in monolingual scene text recognition.

Despite the progress made on the application of visual transformers to STR tasks, it remains an open question how to explicitly identify inter-patch correlations for multilingual scene text recognition containing multi-scale, multi-orientation and low-resolution texts. 

\section{TANGER}
\subsection{Overall Architecture}
  Our goal is to introduce a new vision transformer architecture for handling complex multilingual STR tasks.  The proposed TANGER, as shown in Fig.~\ref{fig2}, consists of a primary transformer, which is a normal vision transform, and a supplementary pyramid transformer with adaptive n-grams embeddings. The primary vision transform focuses on the patch-based image features without considering the associated local patches, in which a dropout layer is adopted to alleviate overfitting. The supplementary pyramid transformer concentrates on effectively extracting visual features from the neighbouring image patches with the help of the adaptive n-grams embedding, allowing us to deal with multi-scale and multi-orientation visual appearances of multilingual texts. Similar to \cite{atienza2021vision}, we use a single model architecture in the primary transformer, while in the supplementary transformer, three layers of pyramid transformer encoders are stacked to control the scale of feature maps in three stages. As shown in Fig.~\ref{fig2}, $F_{2}$, $F_{3}$ and $F_{4}$ and the feature maps extracted from the previous stage and serve as the input into encoder layers $L_1$, $L_2$, and $L_3$, respectively. To reduce the computational complexity of TANGER, the encoder in the primary transformer and all three encoder layers in the supplementary transformer share the same parameters.  
  
 In our implementation, we first divide an input RGB image into non-overlapping patches, and the features of a patch is regarded as a concatenation of the raw pixel RGB values. For example, we use a patch size of $m \times m$, and split a given input image of size $H\times W\times 3$ into $\frac{HW}{m^{2}}$ patches, where $m$, $H$ and $W$ are parameters to be defined. After that, a linear embedding layer is applied on this raw-valued feature to project it onto an arbitrary dimension $C$ for each patch. In this way, we can flexibly pass the embedded patches of different dimensions as well as a position embedding to the input of the two transformers. 

\subsection{Patch-based Adaptive n-grams Embedding}
 
The vision transformer in \cite{atienza2021vision} only considers single image patches as a set of isolated vision words or characters without taking into account the correlation between the features in the neighboring image patches, thus limiting the model's ability to recognize complex scene texts. For obtaining efficient relevant features of the image patches, we treat each image patch as a bag of visual words \cite{yang2007evaluating} to capture the relationship between neighboring patches. To this end, we introduce an adaptive $n$-grams embedding method for determining an optimal $n$ for each image patch.

\begin{figure}
  \centering
   \includegraphics[width=1.0\linewidth]{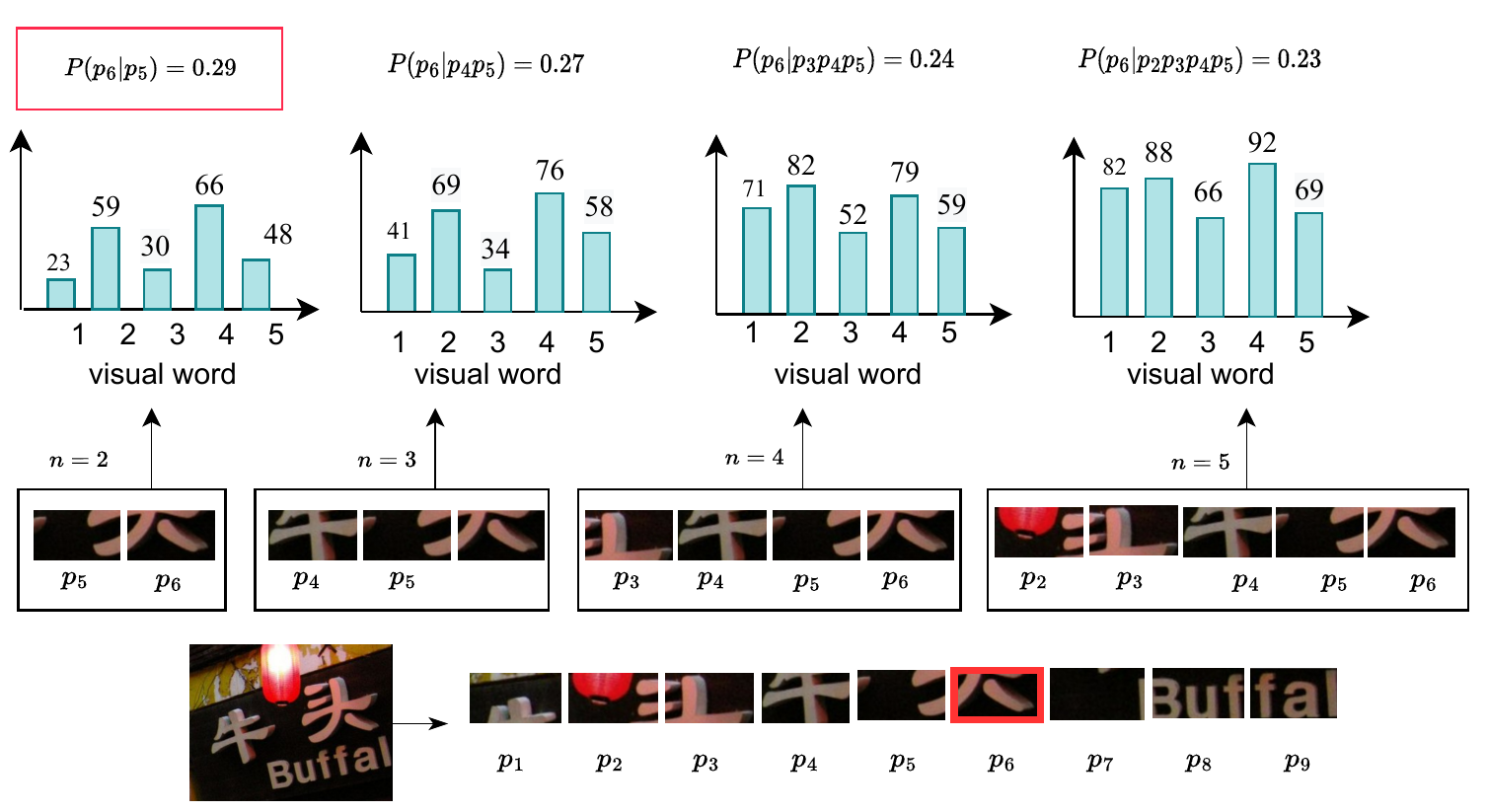}
   \caption{\textbf{An illustrative example of determining the optimal number of grams}. In this example, $p_6$ is the current patch, and bi-gram will be adopted for $p_6$. 
   }
   \label{fig3}
\end{figure}

As illustrated in Fig.~\ref{fig2}, the pyramid transformer architecture \cite{wang2021pyramid} with $n$-grams embedding is employed to flexibly capture the linguistic information in the neighboring patches. However, the value of $n$ for the $i$-th image patch $p_{i}$ must be determined individually to best capture the correlations between its neighboring patches. In this work, we set $n=\{2,3,4,5\}$ according to our pilot studies. Then, we choose the optimal $n_i$ for $p_i$ as follows:
\begin{equation}
  n_i(p_{i}) = \max_{n}{P(p_{i}|p_{i-n+1}p_{i-n+2}...p_{i-1})},n=\{2,3,4,5\},
  \label{eq1}
\end{equation}
 where $P(.)$ is the probability at which $p_i$ is correlated with its previous $(n-1)$ patches, $p_{i-n+1}p_{i-n+2}...p_{i-1}$ represents $n-1$ sequential local visual patches. We can estimate $P(.)$ using a feature histogram of a group of continuous patches built by using the bag of visual words (BVW) as suggested in \cite{yang2007evaluating,tripathi2022bag}: 
\begin{equation}
  {P(p_{i}|p_{i-n+1}p_{i-n+2}...p_{i-1})} = \frac{\max_{k=1}^K\{N(VW_{k})\}}{\sum_{k=1}^{K}N(VW_{k})},
  \label{eq2}
\end{equation}
where $VW_{k}$ is the $k$-th visual word, $K$ is the total number of visual words in the histogram, and $N(VW_{k})$ refers to the frequency of features of the input image patches $p_{i-n+1}p_{i-n+2}...p_{i-1}p_{i}$ that is similar to each of the visual word in the histogram. An illustrative example is given in Fig.~\ref{fig3}, in which the input image is partitioned into nine patches. To determine the number of grams for $p_6$, we calculate $N(VK_k), k=1,\cdots, 5$ when $n=2,3,4,5$, respectively. Finally, the maximum probability is obtained when $n=2$. Consequently, bi-gram embedding is adopted for $p_6$.   
Since the length of the local patches differs from patch to patch (in the above example, the length varies from 2 to 5), we use a spatial pyramid pooling layer to obtain the final n-grams embedding representation that conforms to the input size of the transformer. This way, we can flexibly adjust the length of the local image patches most relevant to the current patch, enabling it to extract the most effective features for complex multilingual scene texts. 

\subsection{Cross-language Rectification Loss}
For multilingual scene text recognition, we must pay particular attention to the variability of the extracted features of two transformers for different languages. To achieve this, we design a loss function by including an extra cross-language rectification loss in addition to the vision loss:   
 \begin{equation}
  Loss_{total} = Loss_{v} + \alpha Loss_{clr},
  \label{eq3}
\end{equation}
where $Loss_{v}$ means the loss for the vision model, which is set following \cite{atienza2021vision}, $Loss_{clr}$ represents the cross-language rectification loss, and $\alpha$ is the weight coefficient, which is set to 0.01 based on our pilot studies. $Loss_{clr}$ is defined as follows:
\begin{equation}
  Loss_{clr} = L_{class} + L_{score},
  \label{eq4}
\end{equation}
where $L_{score}$ is the loss of the word coherence scores, and $L_{class}$ is the loss of language class identification:
\begin{equation}
  L_{class} = SoftCE(Maxpool(MLP(T_{pt}) + MLP(T_{st}))).
  \label{eq5}
\end{equation}
where $MLP(\cdot)$ is a multilayer perceptron (MLP) in the transformers that performs feature extraction, and $T_{pt}$ and $T_{st}$ denote the extracted cross-language features from the vision transformer and pyramid transformer, respectively, $SoftCE(.)$ is the soft cross entropy of language class. With the help of the supplementary transformer, the coherence scores of the predicted characters in different patches are obtained by the linear inference layer in MLP, and the loss of the word coherence scores for multilingual texts can be estimated as follow:  
 \begin{equation}
    L_{score} = -log (Linear (\arg max(y_{i}))),
  \label{eq7}
 \end{equation}
where $y_{i}, i \in [1, maxlen]$ represent inference result of the $i$-th character, $maxlen$ is the maximum length of the predicted characters. 

\subsection{Discussion}
Here, we discuss further the relationship between TANGER and ViTSTR \cite{atienza2021vision}. Like ViTSTR,  TANGER is an image-based transformer model without relying on language resources such as dictionaries or corpus. Similar to the traditional vision transformer \cite{dosovitskiy2020image}, ViTSTR is mainly concerned with extracting features from a sequence of separate image patches. By contrast, TANGER explicitly takes into account the potential correlation between neighbouring visual patches, which is of paramount importance for feature extraction in complex scenes, such as multi-scale, multi-orientation and multilingual texts \cite{nanda2019illumination}. 


\section{Experimental Results}
In this section, we experimentally validate our proposed TANGER by comparing the performance with the state-of-the-art methods on several public datasets as well as one newly collected multilingual dataset TsiText. First, we examine the performance of TANGER for multilingual scene text recognition in comparison with two end-to-end methods \cite{buvsta2018e2e, huang2021multiplexed} and one dictionary-guided method \cite{nguyen2021dictionary}. Then, we compare our model with the vision transformer ViTSTR \cite{atienza2021vision} in three variants, \textit{i.e.}, tiny, small, and base versions for monolingual scene text recognition. Finally, we present the results of ablation studies comparing TANGER with its two variants, one without the adaptive $n$-gram patch embedding method, and the other without cross-language rectification for multilingual scene text recognition. 

\subsection{Experimental Settings}
\textbf{Datasets}\quad
We validate the effectiveness of TANGER on three multilingual datasets, MLT17 \cite{buvsta2018e2e}, CTW1500 \cite{yuliang2017detecting}, RCTW17 \cite{shi2017icdar2017}, one monolingual dataset Totaltext \cite{ch2017total}, as well as a new multilingual dataset  TsiText. MLT2017 \cite{saha2020multi}, which comes from ICDAR 2017 Robust Reading Competition, contains 7200 training, 1800 validation and 9000 testing natural scene images in Arabic, Latin, Chinese, Japanese, Korean, and Bangla. CTW1500 \cite{yuliang2017detecting} includes 1000 natural scene images for training and 500 for testing in English and Chinese languages. RCTW17 \cite{shi2017icdar2017} contains 8034 training and 4229 testing images, including primarily on-scene texts in Chinese and a few in English. Totaltext \cite{ch2017total} contains 1255 training images and 300 images for testing with 11459 annotated
text instances of wild scenes in English.  

TsiText contains 3600 training, 600 validation, and 1800 testing natural images collected from tourism scenes. All images in TsiText are downloaded from the Internet and do not contain sensitive information. There are an average of four text instances in per scene image, with a maximum of 61 text instances. Each text instance contains four different attributes, including position, character content, scene class, and language, and is annotated in a similar way to \cite{yuliang2017detecting}. TsiText aims to provide complex multilingual scene texts containing rich and diverse travel-related scenes such as food, shop signs, traffic signs, directions, landmarks, and posters. It also includes multiple text types, including handwriting, print, and complex artistic styles. Finally, there is certain degrees of variety in text shapes, such as horizontal, multi-directional, curved, circular, and partially obscured texts, among others. To the best of our knowledge, this is the richest dataset for multilingual tourism scene texts. 

\textbf{Metrics and Parameter Configurations} \quad 
We adopt the character-level accuracy as the metric for comparing the algorithms on multilingual scene text recognition tasks. In addition, the frequency of the word-level edit distance between the predicted and ground truth words is employed to compare the recognition performance of the algorithms under comparison on Latin and non-Latin language scene texts. To verify the effectiveness of TANGER on monolingual scene texts, we compare it with ViTSTR in terms of model parameters, speed, FLOPS as well as accuracy.

During the training, we adopt Adam with a mini-batch size of 192 to train the transformer models for 300 epochs from scratch with a learning rate of 0.001 on two A100 GPUs. 

\subsection{Performance on Multilingual Scene Texts}
Table \ref{tab1} lists the performance of the proposed TANGER with three state-of-the-art algorithms for multilingual scene text recognition, namely ABCNet+D \cite{nguyen2021dictionary}, E2E-MLT \cite{buvsta2018e2e}, and Multiplexed \cite{huang2021multiplexed}. ABCNet+D \cite{nguyen2021dictionary} is a dictionary-based recognition algorithm that incorporates dictionaries before handling ambiguous cases. E2E-MLT \cite{buvsta2018e2e} is an end-to-end approach with a single fully convolutional network applicable to both Latin and non-Latin languages for multilingual scene text. Multiplexed \cite{huang2021multiplexed} proposes a unified loss with disentangled loss and integrated loss, and trains multiple text recognition heads in an end-to-end manner for script identification in different languages. Note that E2E-MLT and Multiplexed are end-to-end methods for both text detection and recognition, however, this work focuses on the recognition performance of the compared algorithms.

As listed in Table \ref{tab1}, we can see that TANGER achieves the best character-level accuracy on four multilingual scene text recognition tasks. Compared with ABCNet+D, a dictionary-guided approach, TANGER can significantly improve the accuracy by 14.8\%, 11.9\%, 10.2\%, and 10.1\% on MLT17, LSVT19, RCTW17, and TsiText datasets, respectively.   
The impressive results of TANGER may be attributed to the fact that visual feature extraction is more effective than using lexicons when dealing with a series of multilingual texts in complex scenes. Experimental results also show that TANGER consistently outperforms two end-to-end approaches on all multilingual scene text recognition tasks. Specifically, TANGER achieves a state-of-the-art accuracy of 55.1\% and 89.9\% on the MLT17 and TsiText datasets, respectively. We surmise that the competitive performance of TANGER comes mainly from the more effective representation of the neighboring visual features, as well as the use of cross-language rectification for complex multilingual scene texts, which has also been verified in our ablation studies. 

\begin{figure*}
  \centering
   \includegraphics[width=0.9\linewidth]{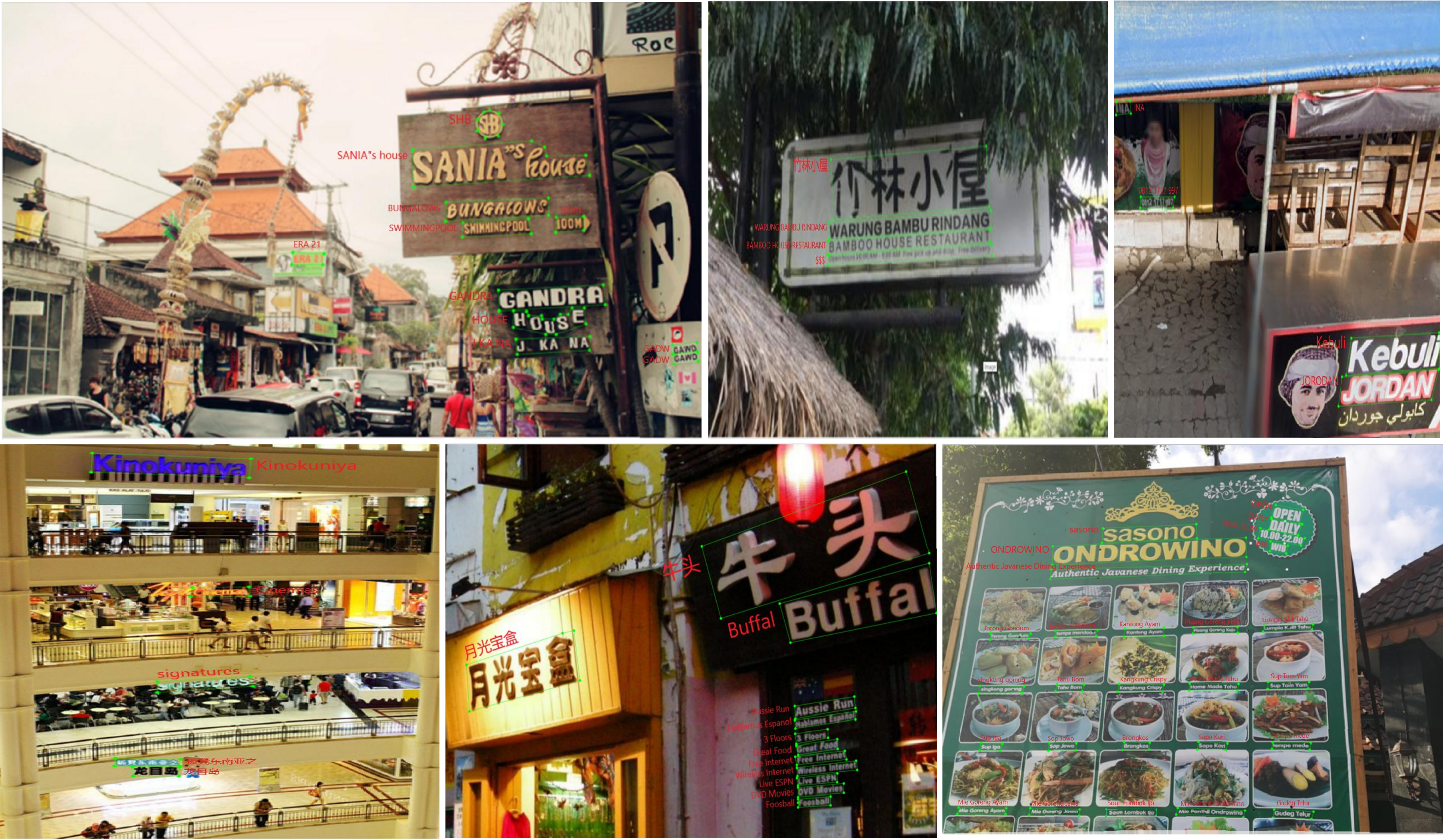}
   \caption{Multilingual text recognition results of TANGER on six complex images from the TsiText dataset. }
   \label{fig4}
\end{figure*}

Figure \ref{fig4} exemplifies the recognition results by TANGER on six complex testing images from the TsiText dataset. We see that TANGER can successfully recognize the texts in multiple languages in various complex scenes. On these six images, TANGER achieves an accuracy of 99.1\%, while ABCNet+D, E2E-MLT, and Multiplexed achieve 91.2\%, 94.6\% and 95.1\%, respectively, confirming the benefits of the proposed adaptive n-grams embedding and cross-language rectification mechanisms.     

\begin{figure}
  \centering
   \includegraphics[width=1.0
   \linewidth]{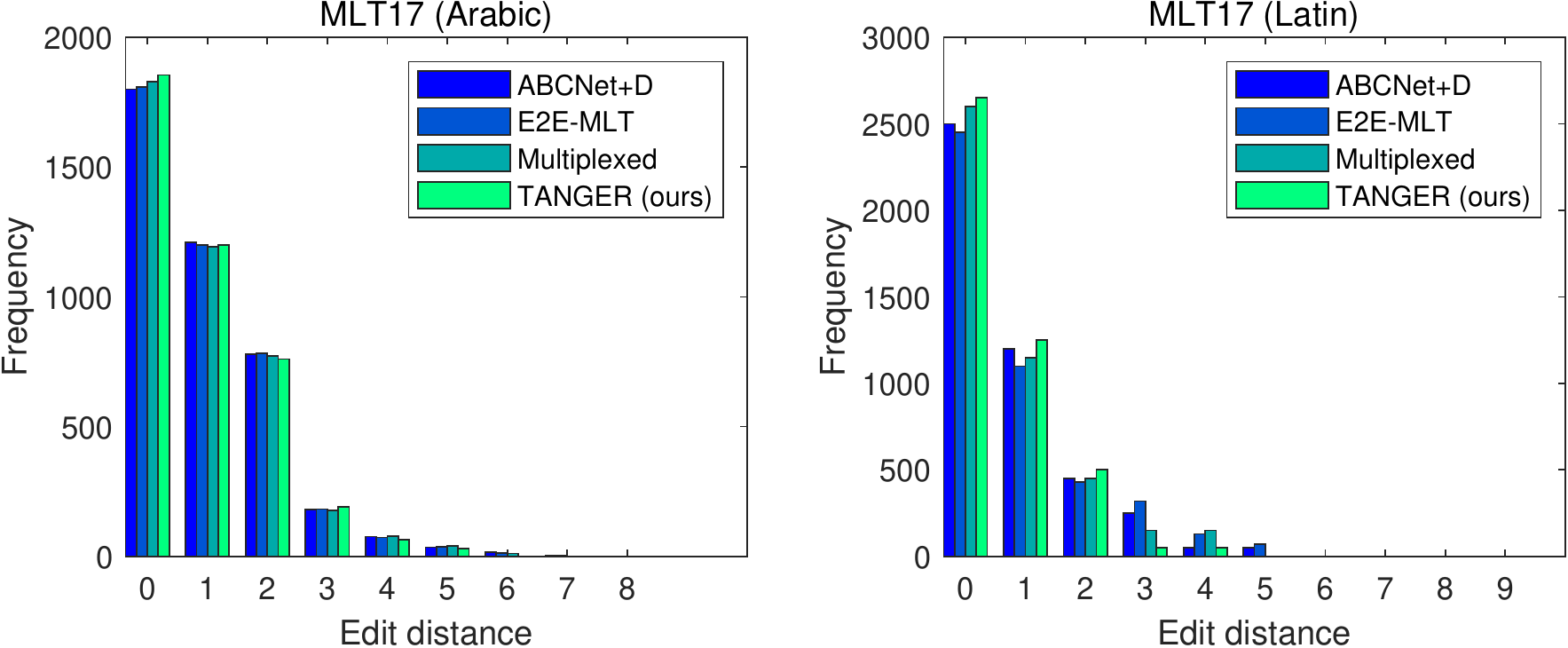}
   \caption{Histogram of word frequencies on different edit distances for Arabic (non-Latin) and Latin language text on MLT17.}
   \label{fig5}
 \end{figure}

To further demonstrate the performance of TANGER for multilingual scene text recognition on both Latin and non-Latin languages, we examine the histograms of edit distance \cite{saluja2017error} between the pairs of predicted and the ground truth words for Arabic (non-Latin) and Latin language texts on the MLT17 dataset. Note that the smaller the edit distance, the smaller the recognition error of the algorithm is, and the words are recognized correctly when the edit distance is 0. From Fig.~\ref{fig5}, we observe that TANGER can obtain the maximum frequency of occurrences at an edit distance of 0 among all compared methods for both Arabic and Latin languages. In addition, the edit distance of TANGER is always smaller than 5 for Latin languages. Overall, we can conclude that TANGER outperforms the compared methods by observing the histograms of the edit distance, further demonstrating the effectiveness of TANGER for multilingual scene text recognition.

\begin{table}
\small
  \centering
  \caption{Recognition results in terms of character-based accuracy of four compared methods on four multilingual scene text datasets. The  character-level accuracy is adopted as the metric for recognition qualities for multilingual scene text recognition.}
  \begin{tabular}{cccccc}    
  \toprule
    Method & MLT17 &  CTW1500 & RCTW17 &   TsiText \\
    \midrule
    ABCNet+D \cite{nguyen2021dictionary}& 40.3 &  73.3 & 63.2 & 79.8\\
    E2E-MLT \cite{buvsta2018e2e}& 49.1 & 65.3 & 69.2  & 84.9\\ 
    Multiplexed \cite{huang2021multiplexed}& 53.2 &  80.3 & 72.2 & 88.3 \\
    TANGER(ours)& \textbf{55.1} &  \textbf{85.2} &  \textbf{73.4} & \textbf{89.9}\\
    \bottomrule
  \end{tabular}
  \label{tab1}
\end{table}

\subsection{Performance on Monolingual Scene Texts}
  Here we evaluate TANGER for monolingual scene text recognition on the Totaltext dataset by comparing it with ViTSTR \cite{atienza2021vision}. Table \ref{tab2} reports the comparative results in terms of the accuracy, speed, model parameters, and FLOPS for three variants of the two compared models, tiny, small, and base versions. 
  Since the transformer-based ViTSTR model is limited to Latin language text recognition 
  we compared the two algorithms on Totaltext, which contains English texts only. From the results in Table \ref{tab2}, we can find that TANGER enhances the recognition accuracies by 2.1\%, 3.6\% and 1.1\% for the tiny, small and base versions, respectively, compared to ViTSTR, without considerably slowing down the inference speed.
  
\begin{figure}
  \centering
   \includegraphics[width=0.95\linewidth]{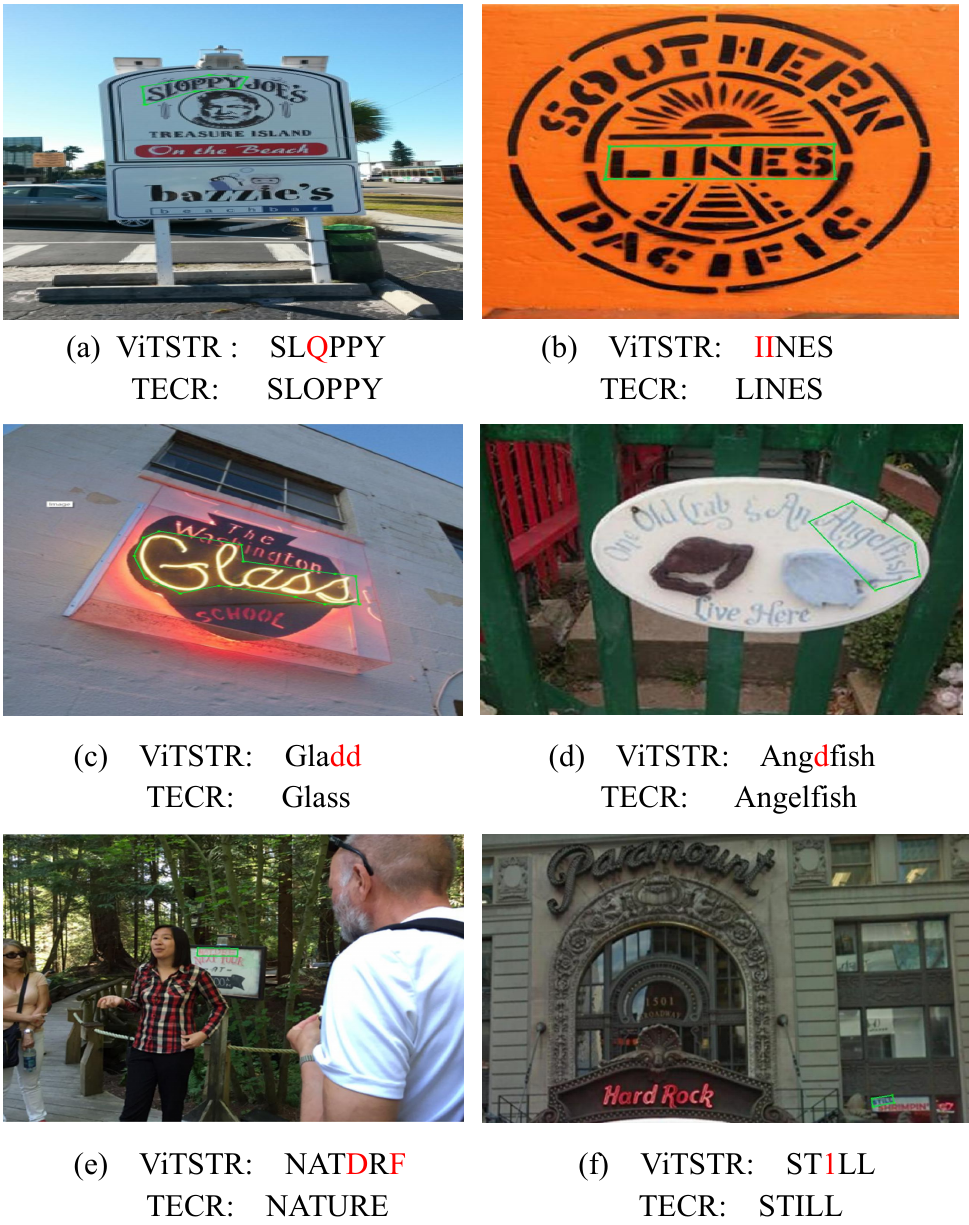}
   \caption{Six selected cases representing typical complex monolingual text scenarios, where ViTSTR makes mistakes while TANGER does not.}
   \label{fig6}
\end{figure}

 Interestingly, we find that the coherence score loss designed for cross-language rectification in our method is also helpful for enhancing the recognition performance for monolingual tasks, which may be attributed to the supplementary transformer that takes neighboring patches into account for complex scene text extraction. Figure~\ref{fig6} provides some selected cases, where ViTSTR makes mistakes whilst TANGER correctly recognizes the scene texts in some artistic fonts, or multi-scale and multi-orientation texts. 
 

\begin{table}
\small
  \centering
  \caption{Experimental results comparing the recognition accuracy, speed, model parameters, and FLOPS. Three configurations of TANGER and ViTSTR are compared, namely tiny, small, and base version as used in \cite{atienza2021vision}.} 
  \begin{tabular}{lccccc}
    \toprule
    
    \multirow{2}{*}{Method} & Acc & Speed & Parameters                & FLOPS \\
                        & $\% $      & msec/image & $1 \times 10^{6}$ & $1\times 10^{9}$     \\
    \midrule
    ViTSTR-Tiny  & 80.6 &  8.1 & 5.2 & 1.6\\
    ViTSTR-Small & 81.3 & 8.5 & 21.2  & 4.6\\ 
    ViTSTR-Base& 84.9 &  9.1 & 83.7 & 17.3 \\ \hline
    TANGER-Tiny & 82.3 & 8.1 & 5.2  & 1.7\\ 
    TANGER-Small& 84.2 &  9.2 & 22.4 & 5.1 \\
    TANGER-Base& 85.8 &  9.5 &  84.6 & 18.2\\
    \bottomrule
  \end{tabular}  
  \label{tab2}
\end{table}

\subsection{Ablation Studies}
In this section, we conduct ablation studies on MLT17 and TsiText datasets. Our goal is two-fold. First, to assess the importance of the proposed adaptive n-grams embeddings for multilingual text representation, we adopt different but fixed values of $n$ on local neighboring image patches for multilingual text representation compared to the proposed adaptive n-grams embeddings in TANGER. Then, we examine the effect of the proposed cross-linguistic loss function on TANGER's recognition performance.
  
\textbf{Adaptive n-grams Patch Embedding}
The proposed adaptive n-grams embedding method aims to choose an optimal $n$ for each image patch. To demonstrate its benefit, we compare the performance of TANGER variants when $n$ is set to $n=2,3,4,5$, respectively. The comparative results are given Table \ref{tab3}, from which we can clearly see that the adaptive n-grams embedding has led to significant improvements in the performance for multilingual scene text recognition on both MLT17 and TsiText. Specifically, the adaptive n-grams embedding can achieve a maximum performance increase of 1.9\% and 3.2\% on MLT17 and TsiText, respectively, compared with the best variant with a fixed $n$. A possible explanation is that the font sizes of different scene texts in different languages may differ dramatically, which requires an adaptive setting of the n-grams embedding. 

\begin{table}
 \small
  \centering
  \caption{Comparison of TANGER variants with different but fixed $n$ in n-grams embedding on MLT17 and TsiText.} 
  \begin{tabular}{lcccc}
    \toprule
Dataset                  & \quad Method     & \quad Accuracy \\
  \midrule
\multirow{5}{*}{MLT17}   & \quad TANGER(n=2)  &   46.3         \\
                         & \quad TANGER(n=3)  &   48.6            \\
                         & \quad TANGER(n=4)  &   53.2            \\
                         & \quad TANGER(n=5 )  &   52.8             \\
                         & \quad \textbf{TANGER(ours)} &   \textbf{55.1}        \\ \hline
\multirow{5}{*}{TsiText} & \quad TANGER(n=2)  &    84.7            \\
                         & \quad TANGER(n=3)  &    85.6           \\
                         & \quad TANGER(n=4)  &    86.7            \\
                         & \quad TANGER(n=5)  &    85.9            \\
                         & \quad \textbf{TANGER(ours)}  &   \textbf{89.9}         \\
\bottomrule
\end{tabular}
 
  \label{tab3}
\end{table}

\textbf{Cross-language Rectification} \quad 
Next we consider the effect of the proposed cross-language loss functions on the performance of TANGER on the MLT17 dataset. We consider two metrics for comparison, including the character-based accuracy and the frequencies of the word-level edit distance. From the results in Table \ref{tab4}, we see that the accuracy of TANGER has an increase of 3.5\% on MLT17 and 2.3\% on TsiText with cross-language rectification loss. These results indicate that the cross-language loss is able to improve recognition performance on multilingual scenes.

\begin{table}
\small
  \centering
  \caption{Results of TANGER on multilingual scene text recognition on MLT17 with or without the cross-language loss function.} 
\begin{tabular}{lcc}
    \toprule
\multirow{2}{*}{Dataset} & \multicolumn{2}{l}{\begin{tabular}[c]{@{}l@{}}Cross-language rectification used? \end{tabular}} \\
                        & No                                              & Yes                                             \\
                         \midrule
MLT17                      &  51.6                                               & \textbf{55.1} \\
TsiText                    &  87.6                                               & \textbf{89.9}\\
\bottomrule
\end{tabular}
  \label{tab4}
\end{table}

\section{Conclusions}
We have proposed a novel transformer-based architecture with adaptive n-grams embeddings and cross-language rectification to tackle complex multilingual scene text recognition. To the best of our knowledge, this is the first transformer-based approach to multilingual scene text recognition. TANGER can leverage the local neighboring visual patches with the help of the proposed adaptive n-grams embedding method, which is highly beneficial for handling complex scene texts containing multi-scale and multi-orientation texts. In addition, a cross-language loss function is suggested on the basis of a primary and supplementary transformers to effectively recognize multilingual scene texts. Finally, a new database containing complex multilingual tourism scenes is introduced, providing a challenging benchmark for multilingual scene text recognition. In the future, we plan to extend the proposed method to an end-to-end approach containing text detection. In addition, we will explore the proposed augmented transformer architecture for multi-modal text recognition tasks.     

\section*{Acknowledgment}
This work was supported in part by the National Natural Science Foundation of China under Grant No. 62006053 and in part by the Program of Science and Technology of Guangzhou under Grant No. 202102020878 and No. 202102080491. Y. Jin is funded by an Alexander von Humboldt Professorship for Artificial Intelligence endowed by the German Federal Ministry of Education and Research.

\bibliographystyle{IEEEtran}
\small
\bibliography{tanger}

\begin{thebibliography}{10}
\providecommand{\url}[1]{#1}
\csname url@samestyle\endcsname
\providecommand{\newblock}{\relax}
\providecommand{\bibinfo}[2]{#2}
\providecommand{\BIBentrySTDinterwordspacing}{\spaceskip=0pt\relax}
\providecommand{\BIBentryALTinterwordstretchfactor}{4}
\providecommand{\BIBentryALTinterwordspacing}{\spaceskip=\fontdimen2\font plus
\BIBentryALTinterwordstretchfactor\fontdimen3\font minus
  \fontdimen4\font\relax}
\providecommand{\BIBforeignlanguage}[2]{{%
\expandafter\ifx\csname l@#1\endcsname\relax
\typeout{** WARNING: IEEEtran.bst: No hyphenation pattern has been}%
\typeout{** loaded for the language `#1'. Using the pattern for}%
\typeout{** the default language instead.}%
\else
\language=\csname l@#1\endcsname
\fi
#2}}
\providecommand{\BIBdecl}{\relax}
\BIBdecl

\bibitem{han2020survey}
K.~Han, Y.~Wang, H.~Chen, X.~Chen, J.~Guo, Z.~Liu, Y.~Tang, A.~Xiao, C.~Xu,
  Y.~Xu \emph{et~al.}, ``A survey on visual transformer,'' \emph{arXiv preprint
  arXiv:2012.12556}, vol.~2, no.~4, 2020.

\bibitem{zhou2017scene}
B.~Zhou, H.~Zhao, X.~Puig, S.~Fidler, A.~Barriuso, and A.~Torralba, ``Scene
  parsing through ade20k dataset,'' in \emph{Proceedings of the IEEE conference
  on computer vision and pattern recognition}, 2017, pp. 633--641.

\bibitem{liu2021swin}
Z.~Liu, Y.~Lin, Y.~Cao, H.~Hu, Y.~Wei, Z.~Zhang, S.~Lin, and B.~Guo, ``Swin
  transformer: Hierarchical vision transformer using shifted windows,'' in
  \emph{Proceedings of the IEEE/CVF International Conference on Computer
  Vision}, 2021, pp. 10\,012--10\,022.

\bibitem{dosovitskiy2020image}
A.~Dosovitskiy, L.~Beyer, A.~Kolesnikov, D.~Weissenborn, X.~Zhai,
  T.~Unterthiner, M.~Dehghani, M.~Minderer, G.~Heigold, S.~Gelly \emph{et~al.},
  ``An image is worth 16x16 words: Transformers for image recognition at
  scale,'' \emph{arXiv preprint arXiv:2010.11929}, 2020.

\bibitem{atienza2021vision}
R.~Atienza, ``Vision transformer for fast and efficient scene text
  recognition,'' in \emph{International Conference on Document Analysis and
  Recognition}.\hskip 1em plus 0.5em minus 0.4em\relax Springer, 2021, pp.
  319--334.

\bibitem{devlin2018bert}
J.~Devlin, M.-W. Chang, K.~Lee, and K.~Toutanova, ``Bert: Pre-training of deep
  bidirectional transformers for language understanding,'' \emph{arXiv preprint
  arXiv:1810.04805}, 2018.

\bibitem{chen2020generative}
M.~Chen, A.~Radford, R.~Child, J.~Wu, H.~Jun, D.~Luan, and I.~Sutskever,
  ``Generative pretraining from pixels,'' in \emph{International conference on
  machine learning}.\hskip 1em plus 0.5em minus 0.4em\relax PMLR, 2020, pp.
  1691--1703.

\bibitem{feng2020scene}
X.~Feng, H.~Yao, Y.~Qi, J.~Zhang, and S.~Zhang, ``Scene text recognition via
  transformer,'' \emph{arXiv preprint arXiv:2003.08077}, 2020.

\bibitem{zhu2016scene}
Y.~Zhu, C.~Yao, and X.~Bai, ``Scene text detection and recognition: Recent
  advances and future trends,'' \emph{Frontiers of Computer Science}, vol.~10,
  no.~1, pp. 19--36, 2016.

\bibitem{buvsta2018e2e}
M.~Bu{\v{s}}ta, Y.~Patel, and J.~Matas, ``E2e-mlt-an unconstrained end-to-end
  method for multi-language scene text,'' in \emph{Asian conference on computer
  vision}.\hskip 1em plus 0.5em minus 0.4em\relax Springer, 2018, pp. 127--143.

\bibitem{yuliang2017detecting}
L.~Yuliang, J.~Lianwen, Z.~Shuaitao, and Z.~Sheng, ``Detecting curve text in
  the wild: New dataset and new solution,'' \emph{arXiv preprint
  arXiv:1712.02170}, 2017.

\bibitem{shi2017icdar2017}
B.~Shi, C.~Yao, M.~Liao, M.~Yang, P.~Xu, L.~Cui, S.~Belongie, S.~Lu, and
  X.~Bai, ``Icdar2017 competition on reading chinese text in the wild
  (rctw-17),'' in \emph{2017 14th iapr international conference on document
  analysis and recognition (ICDAR)}, vol.~1.\hskip 1em plus 0.5em minus
  0.4em\relax IEEE, 2017, pp. 1429--1434.

\bibitem{ch2017total}
C.~K. Ch'ng and C.~S. Chan, ``Total-text: A comprehensive dataset for scene
  text detection and recognition,'' in \emph{2017 14th IAPR international
  conference on document analysis and recognition (ICDAR)}, vol.~1.\hskip 1em
  plus 0.5em minus 0.4em\relax IEEE, 2017, pp. 935--942.

\bibitem{xie2019aggregation}
Z.~Xie, Y.~Huang, Y.~Zhu, L.~Jin, Y.~Liu, and L.~Xie, ``Aggregation
  cross-entropy for sequence recognition,'' in \emph{Proceedings of the
  IEEE/CVF conference on computer vision and pattern recognition}, 2019, pp.
  6538--6547.

\bibitem{tounsi2018multilingual}
M.~Tounsi, I.~Moalla, F.~Lebourgeois, and A.~M. Alimi, ``Multilingual scene
  character recognition system using sparse auto-encoder for efficient local
  features representation in bag of features,'' \emph{arXiv preprint
  arXiv:1806.07374}, 2018.

\bibitem{yao2014strokelets}
C.~Yao, X.~Bai, B.~Shi, and W.~Liu, ``Strokelets: A learned multi-scale
  representation for scene text recognition,'' in \emph{Proceedings of the IEEE
  conference on computer vision and pattern recognition}, 2014, pp. 4042--4049.

\bibitem{shi2016end}
B.~Shi, X.~Bai, and C.~Yao, ``An end-to-end trainable neural network for
  image-based sequence recognition and its application to scene text
  recognition,'' \emph{IEEE transactions on pattern analysis and machine
  intelligence}, vol.~39, no.~11, pp. 2298--2304, 2016.

\bibitem{li2019show}
H.~Li, P.~Wang, C.~Shen, and G.~Zhang, ``Show, attend and read: A simple and
  strong baseline for irregular text recognition,'' in \emph{Proceedings of the
  AAAI conference on artificial intelligence}, vol.~33, no.~01, 2019, pp.
  8610--8617.

\bibitem{zhan2019esir}
F.~Zhan and S.~Lu, ``Esir: End-to-end scene text recognition via iterative
  image rectification,'' in \emph{Proceedings of the IEEE/CVF Conference on
  Computer Vision and Pattern Recognition}, 2019, pp. 2059--2068.

\bibitem{shi2018aster}
B.~Shi, M.~Yang, X.~Wang, P.~Lyu, C.~Yao, and X.~Bai, ``Aster: An attentional
  scene text recognizer with flexible rectification,'' \emph{IEEE transactions
  on pattern analysis and machine intelligence}, vol.~41, no.~9, pp.
  2035--2048, 2018.

\bibitem{fang2018attention}
S.~Fang, H.~Xie, Z.-J. Zha, N.~Sun, J.~Tan, and Y.~Zhang, ``Attention and
  language ensemble for scene text recognition with convolutional sequence
  modeling,'' in \emph{Proceedings of the 26th ACM international conference on
  Multimedia}, 2018, pp. 248--256.

\bibitem{huang2021multiplexed}
J.~Huang, G.~Pang, R.~Kovvuri, M.~Toh, K.~J. Liang, P.~Krishnan, X.~Yin, and
  T.~Hassner, ``A multiplexed network for end-to-end, multilingual ocr,'' in
  \emph{Proceedings of the IEEE/CVF Conference on Computer Vision and Pattern
  Recognition}, 2021, pp. 4547--4557.

\bibitem{qiao2020seed}
Z.~Qiao, Y.~Zhou, D.~Yang, Y.~Zhou, and W.~Wang, ``Seed: Semantics enhanced
  encoder-decoder framework for scene text recognition,'' in \emph{Proceedings
  of the IEEE/CVF Conference on Computer Vision and Pattern Recognition}, 2020,
  pp. 13\,528--13\,537.

\bibitem{lyu20192d}
P.~Lyu, Z.~Yang, X.~Leng, X.~Wu, R.~Li, and X.~Shen, ``2d attentional irregular
  scene text recognizer,'' \emph{arXiv preprint arXiv:1906.05708}, 2019.

\bibitem{na2021multi}
B.~Na, Y.~Kim, and S.~Park, ``Multi-modal text recognition networks:
  Interactive enhancements between visual and semantic features,'' \emph{arXiv
  preprint arXiv:2111.15263}, 2021.

\bibitem{raisi2020}
Z.~Raisi, M.~A. Naiel, P.~Fieguth, and S.~Wardell, ``{2D} positional
  embedding-based transformer for scene text recognition,'' \emph{Journal of
  Computational Vision and Imaging Systems}, vol.~6, no.~1, pp. 1--4, 2020.

\bibitem{biten2022latr}
A.~F. Biten, R.~Litman, Y.~Xie, S.~Appalaraju, and R.~Manmatha, ``Latr:
  Layout-aware transformer for scene-text vqa,'' in \emph{Proceedings of the
  IEEE/CVF Conference on Computer Vision and Pattern Recognition}, 2022, pp.
  16\,548--16\,558.

\bibitem{tan2022pure}
Y.~L. Tan, A.~W.-K. Kong, and J.-J. Kim, ``Pure transformer with integrated
  experts for scene text recognition,'' in \emph{European Conference on
  Computer Vision}.\hskip 1em plus 0.5em minus 0.4em\relax Springer, 2022, pp.
  481--497.

\bibitem{wang2022petr}
Y.~Wang, H.~Xie, S.~Fang, M.~Xing, J.~Wang, S.~Zhu, and Y.~Zhang, ``Petr:
  Rethinking the capability of transformer-based language model in scene text
  recognition,'' \emph{IEEE Transactions on Image Processing}, vol.~31, pp.
  5585--5598, 2022.

\bibitem{yang2007evaluating}
J.~Yang, Y.-G. Jiang, A.~G. Hauptmann, and C.-W. Ngo, ``Evaluating
  bag-of-visual-words representations in scene classification,'' in
  \emph{Proceedings of the international workshop on Workshop on multimedia
  information retrieval}, 2007, pp. 197--206.

\bibitem{wang2021pyramid}
W.~Wang, E.~Xie, X.~Li, D.-P. Fan, K.~Song, D.~Liang, T.~Lu, P.~Luo, and
  L.~Shao, ``Pyramid vision transformer: A versatile backbone for dense
  prediction without convolutions,'' in \emph{Proceedings of the IEEE/CVF
  International Conference on Computer Vision}, 2021, pp. 568--578.

\bibitem{tripathi2022bag}
S.~Tripathi, S.~K. Singh, and L.~H. Kuan, ``Bag of visual words (bovw) with
  deep features--patch classification model for limited dataset of breast
  tumours,'' \emph{arXiv preprint arXiv:2202.10701}, 2022.

\bibitem{nanda2019illumination}
A.~Nanda, D.~S. Chauhan, P.~K~Sa, and S.~Bakshi, ``Illumination and scale
  invariant relevant visual features with hypergraph-based learning for
  multi-shot person re-identification,'' \emph{Multimedia Tools and
  Applications}, vol.~78, no.~4, pp. 3885--3910, 2019.

\bibitem{nguyen2021dictionary}
N.~Nguyen, T.~Nguyen, V.~Tran, M.-T. Tran, T.~D. Ngo, T.~H. Nguyen, and
  M.~Hoai, ``Dictionary-guided scene text recognition,'' in \emph{Proceedings
  of the IEEE/CVF Conference on Computer Vision and Pattern Recognition}, 2021,
  pp. 7383--7392.

\bibitem{saha2020multi}
S.~Saha, N.~Chakraborty, S.~Kundu, S.~Paul, A.~F. Mollah, S.~Basu, and
  R.~Sarkar, ``Multi-lingual scene text detection and language
  identification,'' \emph{Pattern Recognition Letters}, vol. 138, pp. 16--22,
  2020.

\bibitem{saluja2017error}
R.~Saluja, D.~Adiga, P.~Chaudhuri, G.~Ramakrishnan, and M.~Carman, ``Error
  detection and corrections in indic ocr using lstms,'' in \emph{2017 14th IAPR
  International Conference on Document Analysis and Recognition (ICDAR)},
  vol.~1.\hskip 1em plus 0.5em minus 0.4em\relax IEEE, 2017, pp. 17--22.

\end{thebibliography}

\end{document}